            [2005/11/29 v0.1.4
 CESCG proceedings sample file]

\documentclass{cescg}[2005/11/12] 
\usepackage{graphicx}
\usepackage{float}
\usepackage{xcolor}
\usepackage{soul}

\usepackage{url}
\usepackage{subcaption}
\usepackage{listings}

\usepackage{hyperref}

\title{Design and Development of a Web-based Tool for Inpainting of Dissected Aortae in Angiography Images}

 \author{Alexander Prutsch\thanks{alexander.prutsch@student.tugraz.at}, Antonio Pepe\thanks{antonio.pepe@tugraz.at}, Jan Egger\thanks{egger@icg.tugraz.at}}
 \affiliation{Institute of Computer Graphics and Vision \\ Graz University of Technology  \\ Graz / Austria \bigbreak Computer Algorithms for Medicine Lab \\ Graz / Austria}

\keywords{Aortic Dissection, Inpainting, Web Application, Medical Image Analysis, Cloud}
\begin{document}

\maketitle

\begin{abstract}
Medical imaging is an important tool for the diagnosis and the evaluation of an aortic dissection (AD); a serious condition of the aorta, which could lead to a life-threatening aortic rupture. AD patients need life-long medical monitoring of the aortic enlargement and of the disease progression, subsequent to the diagnosis of the aortic dissection. Since there is a lack of “healthy-dissected” image pairs from medical studies, the application of inpainting techniques offers an alternative source for generating them by doing a virtual regression from dissected aortae to healthy aortae; an indirect way to study the origin of the disease. The proposed inpainting tool combines a neural network, which was trained on the task of inpainting aortic dissections, with an easy-to-use user interface. To achieve this goal, the inpainting tool has been integrated within the 3D medical image viewer of StudierFenster (\textcolor{blue}{\href{www.studierfenster.at}{www.studierfenster.at}}). By designing the tool as a web application, we simplify the usage of the neural network and reduce the initial learning curve.


\end{abstract}


\keywordlist

\section{Introduction}
Since the emergence of deep learning networks for image inpainting \cite{DBLP:journals/corr/abs-1901-00212}, the quality of inpainted images increased remarkably over the course of the past few years. Driven by this leap of quality, inpainting could be used in more and more application areas to complete missing or masked image regions. In particular, an uprising field of application for inpainting is medical imaging. A related example would be the removal of interfering artifacts on medical images caused by dental fillings \cite{imagerestoration}. 

\begin{figure}[H]
\centering
\includegraphics[width=0.78\columnwidth]{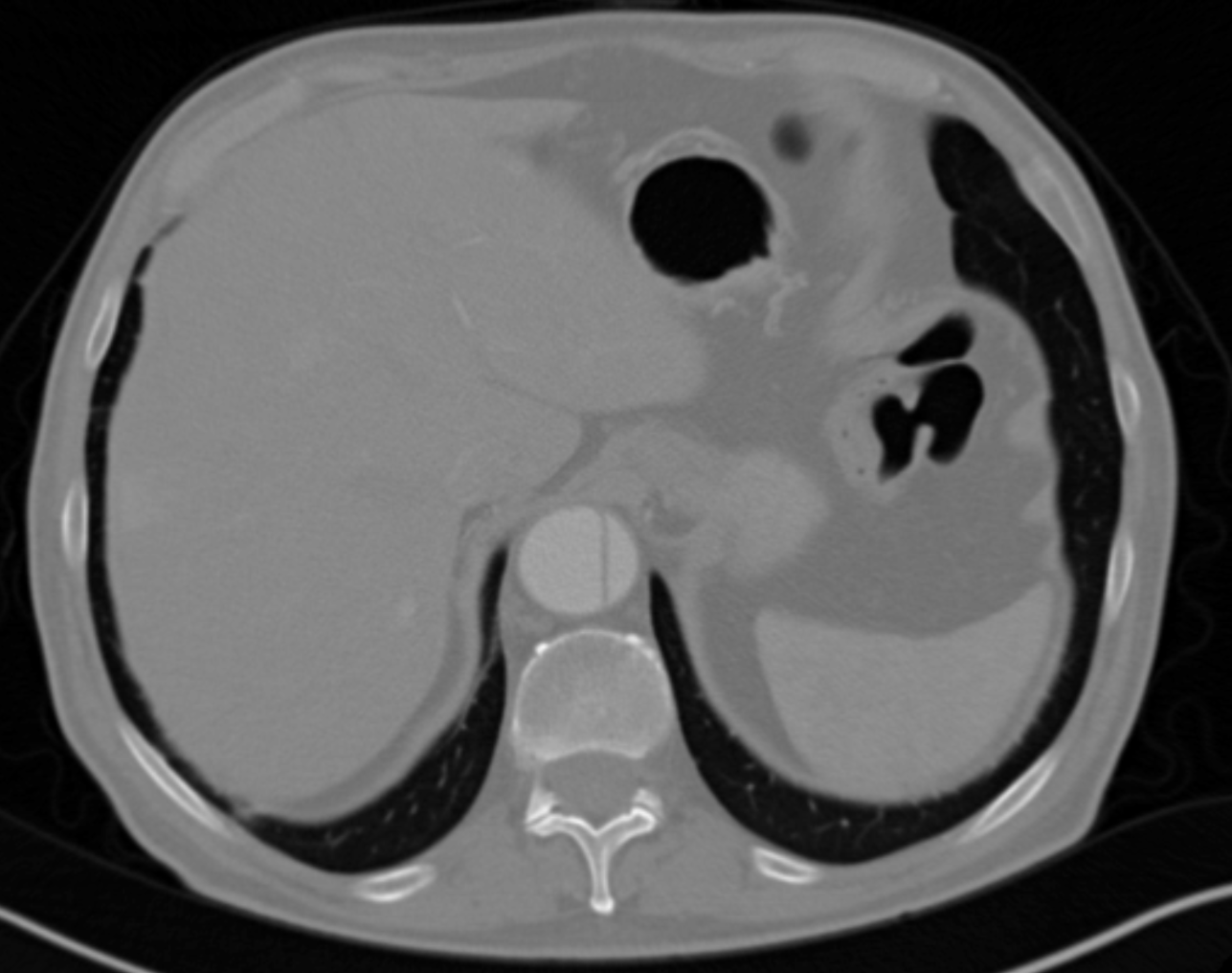}
\caption{Chest CTA scan slice featuring an aortic dissection. The aorta can be recognized as a circular shape in the center of the image (just above the spine). The aortic dissection is visible as a dark line separating this circular shape.}
\label{fig:viewer}
\end{figure}

In this work, we apply inpainting on medical images of aortic dissection. An aortic dissection is characterized by the formation of a second, false lumen in the aorta \cite{AorticDissectionMaps}. The separation of the aortic wall is caused by a tear on the inside of the wall, which allows blood to enter the vascular wall. Due to the dissection of the aortic wall into two parts the structural stability is affected, which could lead to an aortic rupture. Thus, an aortic dissection can evolve into a life-threatening situation and patients usually require a continuous monitoring following the diagnosis of aortic dissection \cite{AorticDissectionMaps}. Medical imaging is important in relation to aortic dissections; its diagnosis is usually based on the interpretation of CTA images \cite{AorticDissectionMaps}. In Figure \ref{fig:viewer} a CTA scan slice showing an aortic dissection can be seen. 

To the best of our knowledge, there are no easily-accessible CTA image pairs available from medical studies, which show a patient before and after the diagnosis of aortic dissection.
Previously, the deep learning model EdgeConnect \cite{DBLP:journals/corr/abs-1901-00212} was trained on inpainting of aortic dissections. EdgeConnect is capable of performing image inpainting using two adversarial models: the first one is used to reconstruct the edges in the missing image regions; the second one is used to complete the missing regions. By executing the neural network, the process can remove the presence of aortic dissection on a given CTA scan slice. Thereby, the visual appearance of the aorta is changed to that of a healthy aorta with a lower cross-sectional diameter. Hence, it is possible to gather image pairs before and with an aortic dissection by simulating the shape of the aorta before the dissection. In future works, these image pairs should help to understand how the shape of the aorta changes and, furthermore, help provide parameters for a more reliable risk assessment. 

EdgeConnect, which is implemented in Python, offers no user interface beside command line interaction. This work describes the integration of the inpainting functionality into a 3D image viewer hosted on a website. As a benefit, the usage of the neural network is simplified and the sparse user experience is enhanced. This website, called StudierFenster (\textit{http://studierfenster.tugraz.at/}), offers different tools for medical image processing. As a result of this work, it is now possible to use the inpainting tool without any software installation directly in the browser. In addition, this integration adds a convenient and simple to understand user interface to the inpainting tool. The user interface features a free-drawing brush for creating the occlusion mask. 

\section{Related Work}
\label{cha:relatedwork}
A similar application of inpainting in the medical imaging area is proposed by Elmahdy et al. \cite{robustcontour}. In this method inpainting is applied on missing regions of CT scans caused by gas pockets in the colon before they are used for image registration. Subsequently, theses images are used for online adaptive proton therapy of prostate cancer.

Additionally, there are also several other web pages with both scientific and commercial purposes, which also offer inpainting functionalities. It should be noted that, to the best of our knowledge, none of them has a focus on medical imaging. An example therefore would be a demonstration tool for DeepFillv2, which is is an generative image inpainting system \cite{yu2018free}. The application supports the creation of a custom mask, but the input image cannot be chosen freely. It is randomly selected out of the CelebA-HQ or places2 dataset as the back-end network is specifically trained on these datasets \cite{deepfill}. Another example would be a commercial showcase tool developed by NVIDIA Corporation. Any image can be uploaded as input image and it also offers a drawing tool for mask creation, which allows the user to draw any arbitrary mask \cite{nvidia}.

\section{Software Components}
In this section, we introduce the core software components of the inpainting tool. These are StudierFenster, the web-based 3D Viewer, and EdgeConnect, the neural network used for image inpainting. 

\subsection{StudierFenster Website}
Studierfenster is a web-based tool for medical visualization developed by researchers from Graz University of Technology and Medical University of Graz \cite{bawild}. In addition to medical image visualization, the website offers tools for data format conversion, image segmentation and the calculation of image scores \cite{segscores}. 

For the front-end, three standard technologies for web development, HTML, JavaScript and CSS, are used in combination with additional JavaScript libraries like JQuery. The back-end is implemented in Python and the core component is a Flask-based application. Flask is a web framework, which means it is capable of handling the communication between a web server and its clients. The functionality of the Flask application is extended by additional back-end CGI modules implemented either in C++ or Python \cite{bawild}. For instance, the tool introduced in this paper embodies one of these modules.


\subsection{Medical 3D Viewer}
The Medical 3D Viewer on StudierFenster allows the user to visualize medical 3D data and execute different operations on the volumetric data such as image labeling \cite{bawild}. Furthermore, the inpainting tool here introduced is developed as extension to the Medical 3D Viewer. The Medical 3D Viewer is based on Slice:Drop \cite{slicedrop}, an interactive viewer for medical imaging data offered by Boston Children's Hospital and Harvard Medical School. Slice:Drop works client-side only and utilizes the X toolkit (XTK) \cite{neuroimagingxtoolkit} for rendering on top of HTML Canvas and WebGL. XTK was also developed by the same team as Slice:Drop in order to provide a lightweight tool for scientific visualisation \cite{neuroimagingxtoolkit}.

\begin{figure}[H]
\centering
\includegraphics[width=1\columnwidth]{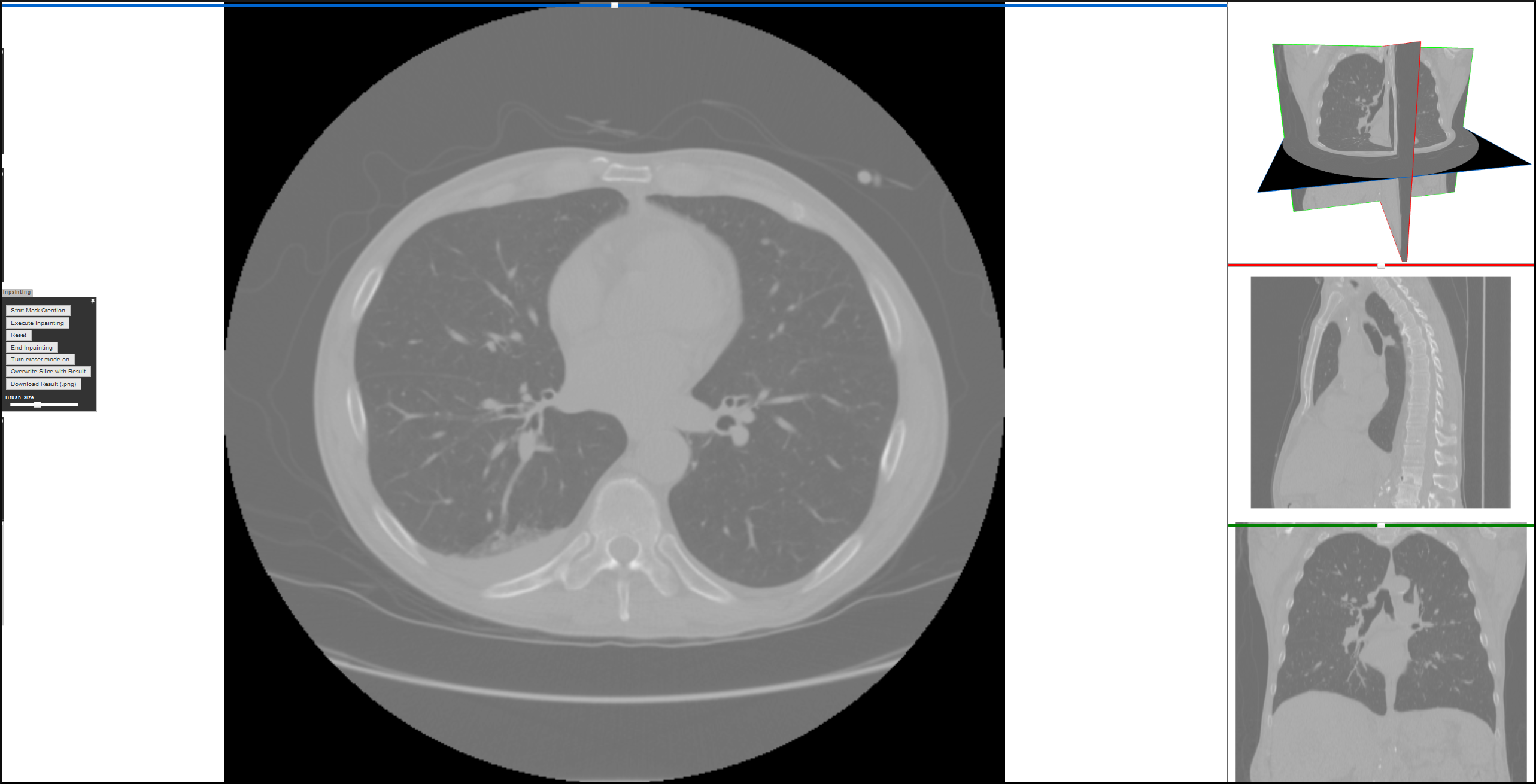}
\caption{Current layout of the Medical 3D Viewer on the StudierFenster website with activated inpainting tab in the sidebar menu (on the left side).}
\label{fig:viewer_layout}
\end{figure}

\begin{figure*}[t]
\centering
\includegraphics[width=2\columnwidth]{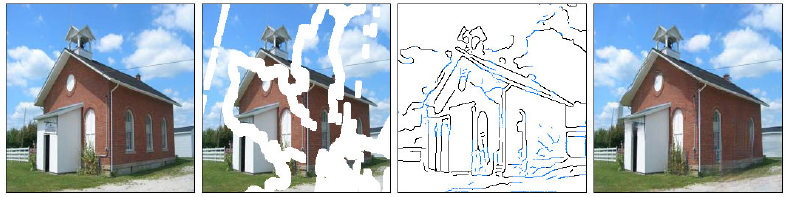}
\caption{An example for the inpainting process using EdgeConnect. From left to right: original image, masked input image, edge image and inpainted image. In the edge image the black edges are detected by using an edge detection algorithm and the blue edges reconstructed by the neural network. Image source: Nazeri et al. \cite{DBLP:journals/corr/abs-1901-00212}.}
\label{fig:edge_connect}
\end{figure*}

Figure \ref{fig:viewer_layout} shows the layout of the Medical 3D Viewer on StudierFenster. The current configuration of the Medical 3D Viewer features four views. Three of them are positioned on the right-side. The fourth one is larger than the others and, hence, is the primary view. In these four views the same CT scan volume is displayed in four different perspectives: a slice in each sagittal plane (from the side), coronal plane (from the front) and axial plane (from the top), as well as a three dimensional rendering, where all three planes are displayed. By clicking on one of the smaller views, the main view can be freely switched. The UI controls for the different operations on input data, like segmentation and inpainting, are located in a sidebar menu.

\subsection{EdgeConnect}
The neural network used for the inpainting tool on the StudierFenster website is based on EdgeConnect \cite{DBLP:journals/corr/abs-1901-00212}, a deep learning model specifically designed for inpainting tasks. EdgeConnect splits the inpainting process into two phases. For both of them two pipe-lined generative adversarial networks \cite{NIPS2014_5423} are used, so overall EdgeConnect is composed of four neural networks: two generators ($G_1$ and $G_2$) and two discriminators ($D_1$ and $D_2$).

The first stage acts as edge completion network; the second stage as image completion network. This means that during the first stage only the missing edges of the input image are reconstructed. The edges of the unmasked region are detected with the Canny edge detection algorithm. In particular, the mask image, the input image as a grey-scale image and the edge map of the unmasked regions are used as input for the first stage. In the second stage the reconstructed edges and the incomplete color image are used to compute a color image where the missing image parts are filled in \cite{DBLP:journals/corr/abs-1901-00212}. 

The images in Figure \ref{fig:edge_connect} illustrate this two-stage workflow. The images show from left to right: the original image, the masked input image (which is an input for both stages), the edge image (which is the result of the first stage and one of the inputs for the second stage) and also the final result (inpainted image) generated with the help of EdgeConnect \cite{DBLP:journals/corr/abs-1901-00212}. 

EdgeConnect is implemented in Python and utilizes the machine learning library PyTorch \cite{DBLP:journals/corr/abs-1901-00212}. For training EdgeConnect on inpainting parts of the aorta, 75 CTA scans of healthy aortae were used. These datasets were taken from the CAD-PE challenge \cite{cadpe} (40 CTA scans) and from Masoudi et al. \cite{masoudi} (35 CTA scans). 

\section{Inpainting Workflow}
\label{cha:workflow}
The inpainting workflow and tool introduced in this work can be divided into different stages. As a prerequisite, the user has to upload a NRRD file containing a dataset of a CT scan, which is subsequently displayed in the Medical 3D Viewer. The rendering of the image data is thereby handled by the Slice:Drop component. Then the user can select the inpainting tool out of several tools for operating on the uploaded dataset.

\begin{figure}[H]
\centering
\includegraphics[width=1\columnwidth]{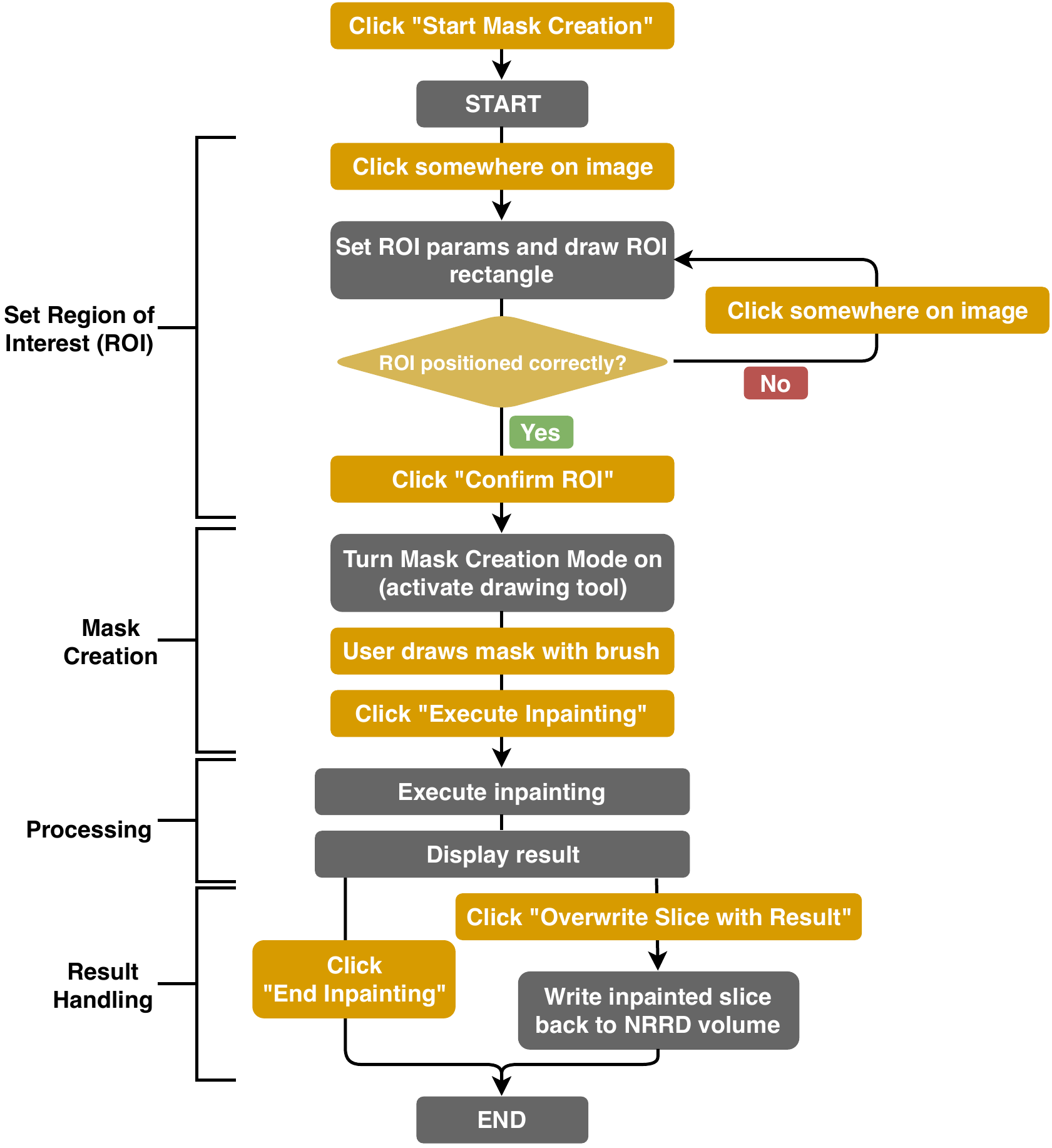}
\caption{Simplified sequence of the inpainting workflow. The orange coloured items mark user inputs, the grey items actions executed by the inpainting tool.}
\label{fig:workflow}
\end{figure}

The stages of the actual process are: definition of a region of interest, drawing a mask, executing the inpainting and handling the result. The following sections will describe these four stages in detail. Everything except the actual execution of the neural network for inpainting is done purely on the client side. Hence, only during the processing stage communication between the client and the server happens. The chart in Figure \ref{fig:workflow} visualizes the principle course of events during the inpainting workflow. Options like resetting the inpainting workflow or cutting it short are not included in this chart. The orange-coloured items mark the interactions of the user with the inpainting tool and the grey-coloured items mark the actions executed by the inpainting tool in order to process the inputs and to perform the inpainting.

\subsection{Set Region of Interest}
For the inpainting task only the image region, which shows the aorta and its surroundings is relevant, but a chest CTA scan usually shows a larger field of view. Additionally, the implementation of the inpainting module in the back-end is only capable of handling input images with a resolution of $100 \times 100$ voxels in the axial plane. Additionally, the implementation of EdgeConnect, which is used for the inpainting tool is only capable of handling input images with a resolution of $100 \times 100$ voxels in the axial plane. But CTA scans have usually a higher resolution than $100 \times 100$ voxels. To gather input data with the required resolution, the user has to set a region of interest (ROI): by clicking on the image the user can place the ROI in form of a rectangle with a predefined size on the relevant image section of the CTA scan. Visually the region of interest is marked by a red rectangle overlaying the input image. The input image and mask drawings outside the region of interest is ignored in all further stages. Once the user has confirmed the position of the region of interest it cannot be moved without a re-initialization of the inpainting workflow.

\subsection{Mask Creation}
To apply the inpainting model on an image, a part of the image has to be masked. The masked part of the input image will be filled by the inpainting model, the remainder serves as context for the model. For creating the mask the user can use a brush, which allows free drawing with a variable brush size. This brush tool was already offered as an extension for the manual segmentation tool \cite{bawild}. For the inpainting tool only adaptions of the functionality of this brush tool were made. For visualization and storage of the mask drawing a half transparent canvas is used. This canvas overlays the canvas, which shows the current slice of the CT scan. The brush tool also features an eraser mode, which can be used to remove parts of the mask by hovering over them.

\subsection{Processing - Execute Inpainting}
\label{cha:processing_execute_inpainting}
After setting up the region of interest and creating a mask, all requirements for executing the inpainting model are fulfilled. Until now, everything was done on the client side; for performance and implementation reasons the inpainting model is executed on the server. Some of the reasons why it is not reasonable to execute it on the client side are, that the execution of the inpainting model is a computation-intensive task, the model definition files are of the order of hundred megabytes large and EdgeConnect is implemented in Python, which cannot be interpreted by commercial browsers.

At the beginning of the execution stage, the data from the input image and the mask underlying the defined region of interest is loaded from the corresponding HTML objects. Afterwards, this data is sent to the server as an AJAX request. AJAX requests are a technique in JavaScript for the communication between a client and a server using the \textit{XMLHttpRequest} JavaScript object. On the server side, the data is stored and the inpainting model script is triggered in a new subprocess. After completion, the resulting image is sent back from the server to the client as a response to the AJAX request.

\subsection{Result Handling}
The inpainting result data returned from the server is placed in a canvas layer on top of the original image. Furthermore, the implementation includes the options to download the inpainted slice as a PNG file or to overwrite the original slice in the NRRD volume with the inpainted slice. Thereafter, the updated volume can be downloaded as a NRRD file by using an already existing function of the Medical 3D Viewer.

\section{Implementation}
\label{cha:implementation}
This section discusses the implementation of the inpainting tool. At first, the client/server architecture and the communication between the front-end and the back-end are described. Afterwards, this section focus on details of the front-end and back-end implementation.

\subsection{Client/Server Architecture}
\label{cha:client_server_architecture}
\begin{figure}[H]
\centering
\includegraphics[width=1\columnwidth]{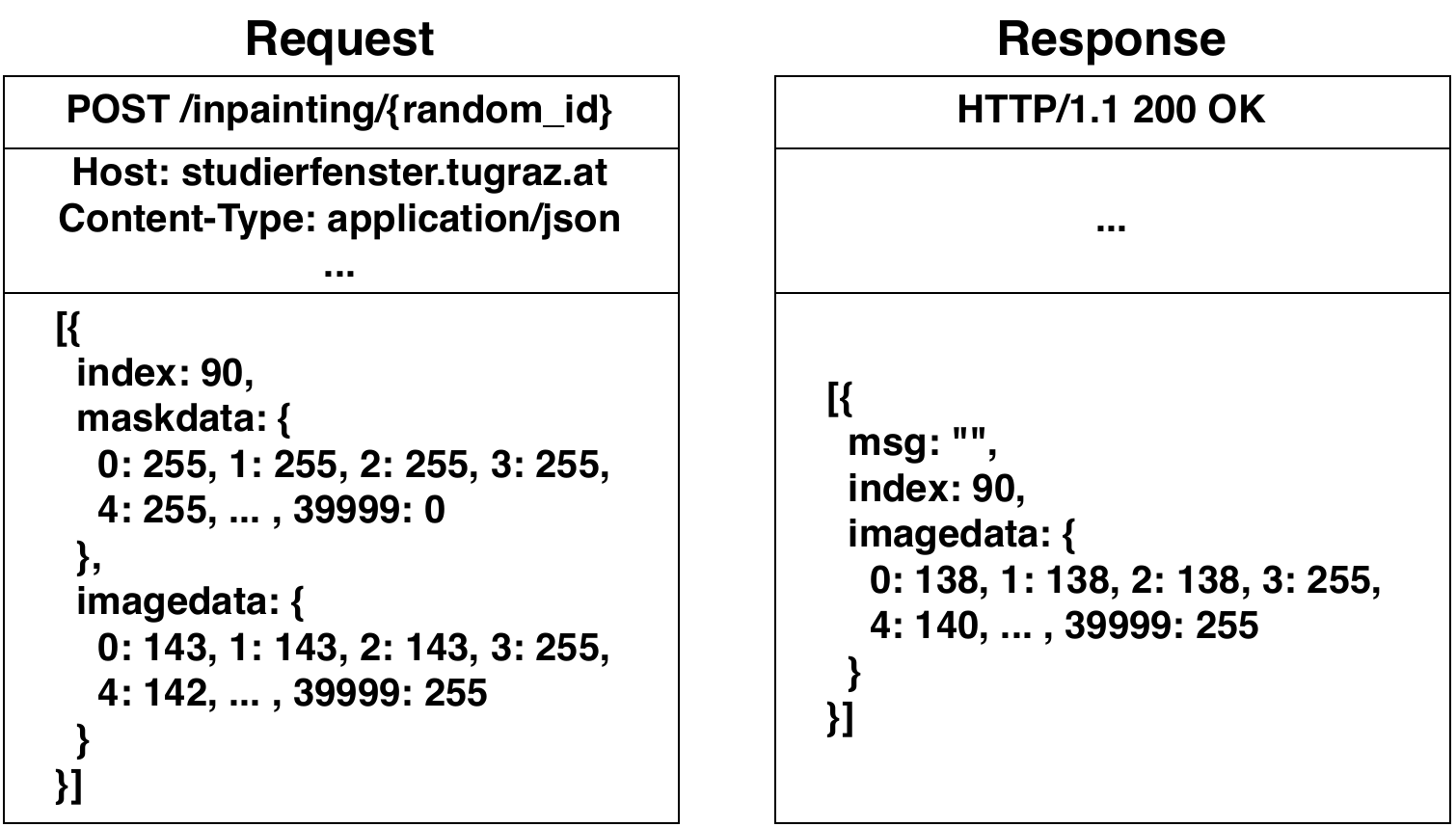}
\caption{Schematic example of the HTTP request and response for sending the input data to the server and receiving the inpainting result.}
\label{fig:req_resp}
\end{figure}

The implementation of the inpainting tool consist of a back-end part, executed on the server, and a front-end part, executed on the client. The communication between both parts consists only of a single HTTP request, which is needed during the processing stage. The request to the server contains a JSON-object with data from the input and mask image. Only the relevant pixels from both images, which are located within the region of interest, are transferred in order to minimize the payload size. In practice, the expected size for the request is of approx. 200 kilobytes. Every incoming request triggers an execution of the inpainting module (EdgeConnect) on the server. The request remains open during the execution and only after the inpainting module has finished the response is sent to the client. On success, the response contains a JSON object with the new image data. If something fails in the back-end, the response contains an error message for debugging purposes. The error message includes the output of EdgeConnect and potential exceptions, which are raised in the Flask application. An example for the content of the HTTP request and response is also given in Figure \ref{fig:req_resp}.


\begin{figure}[H]
\centering
\includegraphics[width=1\columnwidth]{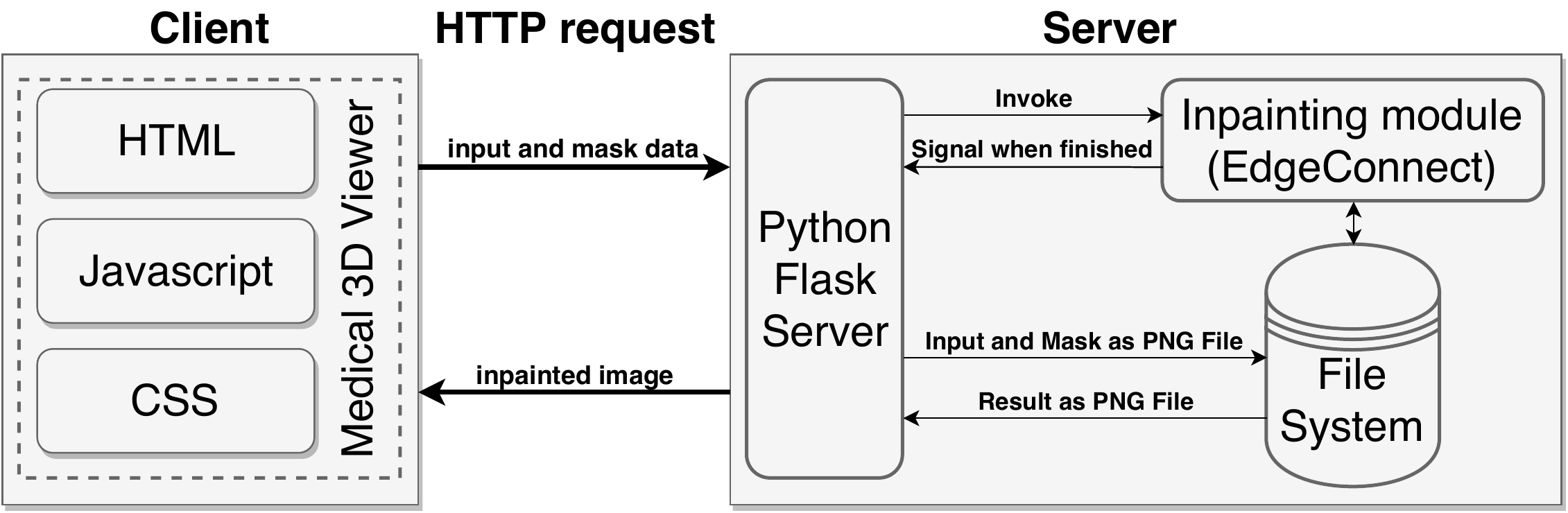}
\caption{Architecture diagram of the inpainting tool (Client-server-model).}
\label{fig:client_server}
\end{figure}

The diagram in Figure \ref{fig:client_server} shows the architecture of the inpainting tool. The separation into front-end and back-end is clearly visible and also the communication between them is illustrated in the diagram. Furthermore, on the back-end side also the communication between the Flask application and the inpainting module is shown, which is realized by read and write operations to the file system (image data) and the Python module \textit{subprocess} (invoke inpainting module and wait for finishing).

\subsection{Front-End}
\label{cha:front-end}
This section discusses the key points of the implementation in the front-end, which are the use of HTML canvases for displaying the graphical elements and the JavaScript code to load the user input and send it to the server. 

\subsubsection{HTML Canvas Element and Canvas Layout}
For image data visualization (CT scan dataset, mask and inpainting result), HTML canvas elements are used. A canvas element is a container for displaying graphical contents. The properties and content of the canvas is defined by using JavaScript code and the RenderingContext object, which corresponds to the canvas element. The RenderingContext object also offers an interface to get or set the image content as an ImageData object. Thereby, the content is represented as a one-dimensional array, which contains the image data line per line. In case of images with multiple channels (like RGBA images), the color channel values for each pixel are stored consecutively. 

\begin{figure}[H]
\centering
\includegraphics[width=1\columnwidth]{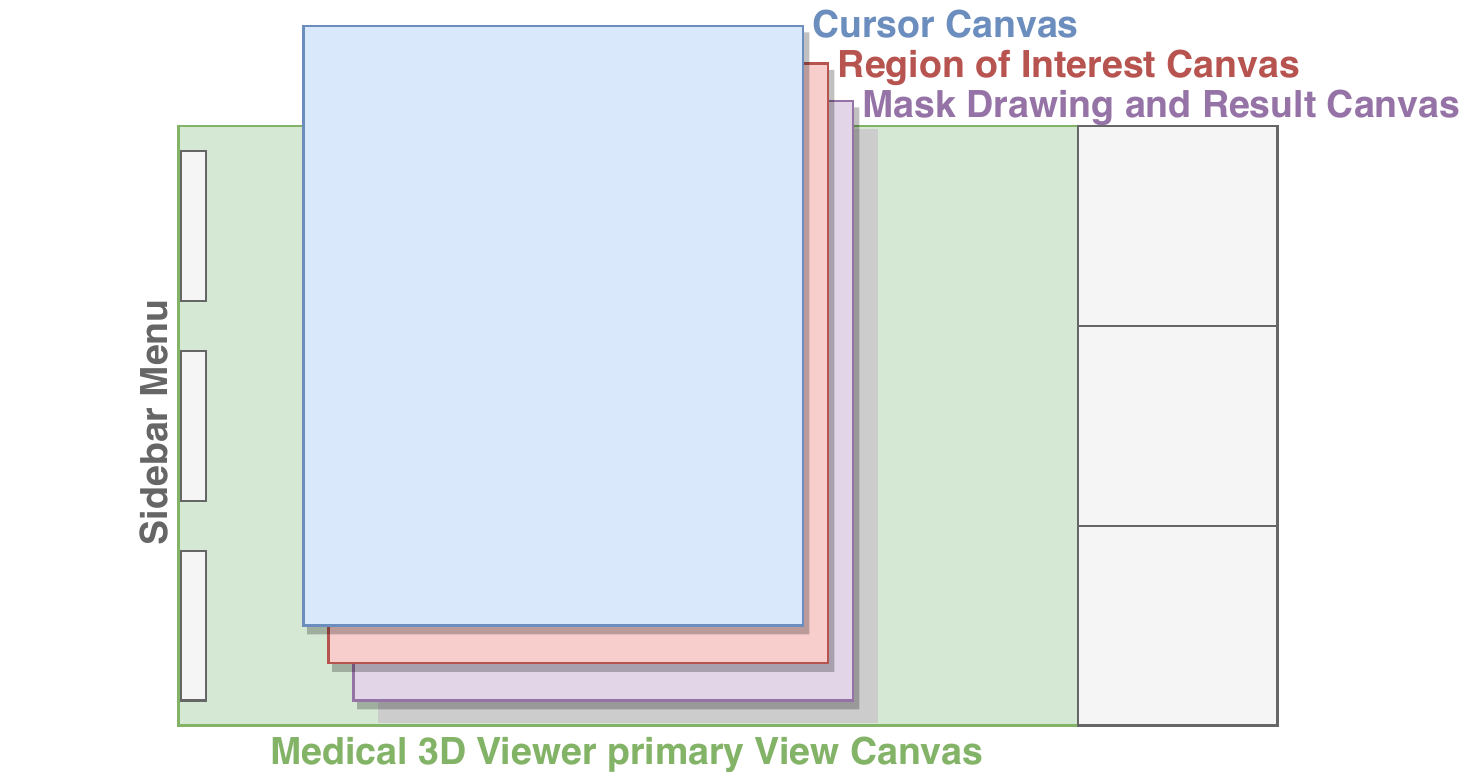}
\caption{Layout of the HTML Canvas elements used for the inpainting tool. See also the screenshot in figure \ref{fig:viewer_layout} for better understanding.}
\label{fig:canvases}
\end{figure}

Similar to other functionalities of the Medical 3D Viewer, like segmentation and landmark detection, it is only possible to use the inpainting tool in the primary view of the Medical 3D Viewer. Furthermore, the tools can only be executed on a slice in axial plane. Slice:Drop renders the file data on a HTML canvas element. To display the region of interest, as well as drawing the mask and display of the result, two more HTML canvas elements are used. These canvas elements are located on top of the primary view canvas. In addition, the free drawing brush tool utilizes another canvas for the visualization of the cursor. Figure \ref{fig:canvases} shows a schematic representation of the HTML canvas layout. 

\begin{figure}[H]
\centering
\includegraphics[width=0.9\columnwidth]{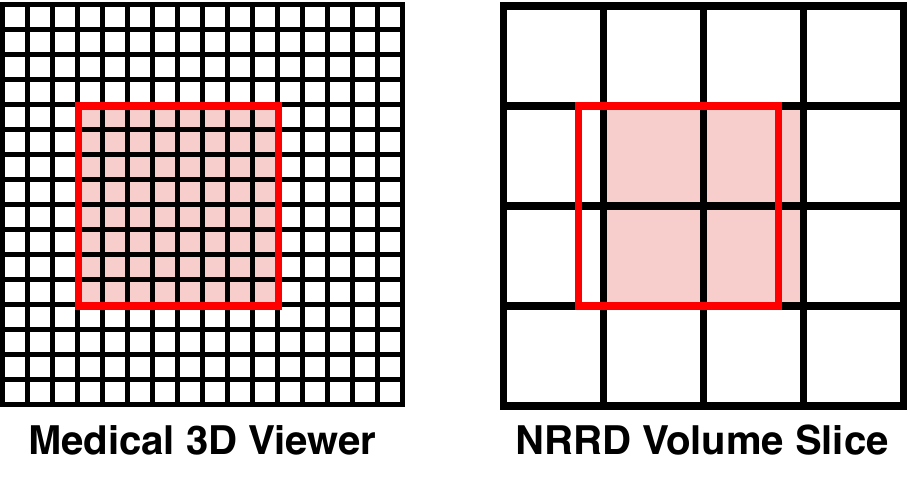}
\caption{Data (inpainting result) displayed in the Medical 3D Viewer (left). Misalignment of data position when writting it back to the lower resoluted NRRD file (right). In practice, the red colored image region in the right sub-figure corresponds to $100 \times 100$ pixels.}
\label{fig:overwritting}
\end{figure}

The earlier mentioned method to get the ImageData is used in order to obtain the input image and mask data from the canvas elements. The method for setting the ImageData is used to put the inpainting result data into a canvas element for visualization. In addition, the HTML Canvas API also offers methods for scaling images, which are also needed for the inpainting tool. When rendering the file data Slice:Drop performs an upscaling to display the CT scans with a higher resolution than the underlying NRRD file data. Otherwise, the CT scans would appear very small. In return, when loading the data of the input image and mask from the canvas elements, it is downscaled to the original resolution. On the other hand, the result of the inpainting module gets upscaled again for display on a HTML canvas. The ratio between the original resolution and the resolution of the rendered image must also be considered for setting the region of interest. To achieve a region of interest with a size of $100 \times 100$ pixels of the original NRRD dataset, the region of interest displayed in the Medical 3D Viewer must be larger by the factor of the scaling ratio. These scaling processes lead to a small displacement of the inpainting result when writing it back to the NRRD file, because the position of the higher resoluted inpainting result does not exactly match the pixel grid of the lower resolved NRRD file (see Figure \ref{fig:overwritting}).

\subsubsection{Data Processing}
After successfully creating a mask, the user can trigger the execution of the inpainting process. Therefore, the main tasks are: loading the input image and the mask image data, send them to the server and handling the result, which is returned from the server. First off, an auxiliary canvas with a size of 100 x 100 pixels is created. Then the stack of slices is scanned with a for-loop, until a slice with a mask drawing is found.

Next, the mask canvas data and input image canvas data need to be downsampled, because, as mentioned earlier, the images in the Medical 3D Viewer are upsampled during the rendering. To accomplish this, the part of the mask canvas, which is within the region of interest, is projected onto the previously mentioned auxiliary canvas using the \textit{drawImageData} function of the HTML Canvas API. Afterwards, the image data of the auxiliary canvas is stored to a new object and the same step is repeated for the input image canvas.

Subsequently, a unique identifier is generated and the object, which holds the mask and input image data is converted to a JSON-object. Then, the data are sent to the server using an AJAX request. If the server successfully returns the image data of the inpainting result, it needs to be put on a HTML canvas element. Therefore, the \textit{drawImageData} function and an auxiliary canvas are used again for upscaling the image data. 

\subsection{Back-End}
\label{cha:back_end}
For the inpainting tool, the Flask application is extended with a new route, which accepts HTTP POST requests. The request contains the data of the input and the mask image as ImageData arrays (which is described above). The unique identifier is also part of the request as an URL parameter.

Both ImageData arrays are then converted into greyscale images. In this case, the reduction from three to one color channel does not lead to a loss of information. Because all color channel values of the input images are equal and for the mask images it is only relevant if a pixel belongs to the mask or not. The reduction is done on the server, because on the client-side an additional iteration over the ImageData arrays would be necessary, while the decrease of the communication overhead would not lead to a evident improvement of the performance. In the same step, the mask image is also converted to a binary mask, which means it only contains two different values. A pixel belongs to the mask, if it contains a part of the drawing done by the user. That implies a pixel is part of the mask if, for instance, the red color channel of the pixel is not equal to 0. Therefore, to create a binary mask, all pixels with a red color channel unequal to 0 are set to 255 (part of mask) and the remaining pixels are set to 0 (not a part of the mask).

Afterwards, these greyscale images can be written to the file system as PNG files. Both files contain a unique identifier in their file names and they have a resolution of $100 \times 100$ pixels. Then the EdgeConnect script is invoked using the Python \textit{subprocess} module. The data transfer between the Flask application and the EdgeConnect process is implemented via the file system in order to keep the required changes to the EdgeConnect source code to a minimum. An analysis of the execution time (see section \ref{cha:results}) also showed, that the file operations only have a small impact on the performance of the inpainting tool, since they are responsible for only a small part of the time needed for processing an inpainting task on the server. The file names of the input image and the mask image are passed to the EdgeConnect process as command line arguments.

The Flask application then waits for the termination of the inpainting process. If it is successful, the result image is processed in opposite manner to the input image and the mask image. First, the PNG file, which contains a RGB image, is read from the file system to a NumPy array and then it is converted to an ImageData object. This ImageData object is then returned to the server as a response to the HTTP POST request. In case of an error of the EdgeConnect process or during the file system operations, only debugging messages are returned to the client. Whether the execution was successful or unsuccessful, in a final step all files, which were created during the handling of that POST request are deleted from the server's file system.

\section{Results}
\label{cha:results}
Following the description of the implementation of the inpainting tool in the previous section, in this section the outcome of this work is described. First, examples of inpainting results are presented. Subsequently, the performance of the inpainting tool is discussed by analyzing the execution time.

\begin{figure} [H]
\centering
\begin{subfigure}[c]{0.32\columnwidth}
\centering
\includegraphics[width=1\columnwidth]{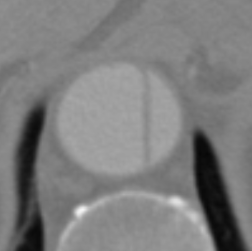}
\caption{Original image}
\label{fig:result_1_original}
\end{subfigure}
\begin{subfigure}[c]{0.32\columnwidth}
\centering
\includegraphics[width=1\columnwidth]{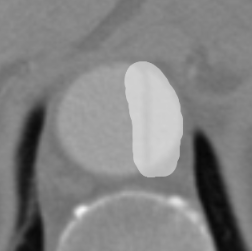}
\caption{Masked image}
\label{fig:result_1_mask}
\end{subfigure}
\begin{subfigure}[c]{0.32\columnwidth}
\centering
\includegraphics[width=1\columnwidth]{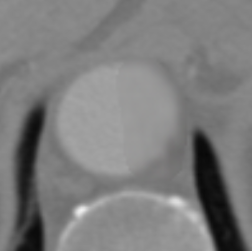}
\caption{Inpainting result}
\label{fig:result_1_result}
\end{subfigure}
\centering
\begin{subfigure}[c]{0.32\columnwidth}
\centering
\includegraphics[width=1\columnwidth]{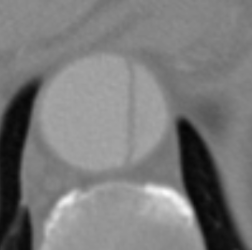}
\caption{Original image}
\label{fig:result_2_original}
\end{subfigure}
\begin{subfigure}[c]{0.32\columnwidth}
\centering
\includegraphics[width=1\columnwidth]{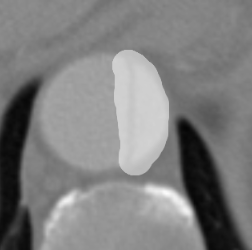}
\caption{Masked image}
\label{fig:result_2_mask}
\end{subfigure}
\begin{subfigure}[c]{0.32\columnwidth}
\centering
\includegraphics[width=1\columnwidth]{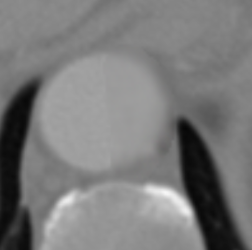}
\caption{Inpainting result}
\label{fig:result_2_result}
\end{subfigure}
\caption{Two examples for an inpainting of an aortic dissection utilizing the inpainting tool.}
\label{fig:result_}
\end{figure}

Figure \ref{fig:result_} shows axial CTA images of an aortic dissection case. In both rows of this figure an example for executing an inpainting with the inpainting tool is given. The image section of all sub-figures is equal to the defined region of interest. The sub-figures in Figure \ref{fig:result_} show from left to right: unedited CTA scan showing the aorta and surroundings, the mask used for the inpainting and the inpainting result. The Sub-images (a) and (b) include a dissected aorta, whereas the dissection is removed in the sub-images (d) and (f).

Only for the processing part of the inpainting workflow (see section \ref{cha:processing_execute_inpainting}) a sensible timing analysis can be done as the other stages are user-dependent, for instance, how much time is spent on drawing the mask. The start of the processing phase is marked by a click on the "Execute Inpainting" button and it is completed after the result is displayed in the Medical 3D Viewer. Furthermore, it can be broken down into five parts: preparing the input data and creating the request (client-side), sending it to the server (network), server-side part, send it back to the client (network) and handling the result (client-side).

At first, the time consumption on the client-side before sending the server request and after receiving the response was evaluated. This was done by calculating the time difference between timestamps logged at four events during the code execution: click on "Execute Inpainting" button, sending the AJAX request, receiving the response to the request and completion of the processing function (see also section \ref{cha:front-end}). Different tests using two browsers (Google Chrome and Mozilla Firefox) showed that the code execution on the client-side only takes 40 to 60 milliseconds.

For measuring the time needed to fulfill the request to the server the network monitor tool of the Mozilla Firefox browser was used. The results show that it takes between 3.7 and 3.9 seconds to fulfill the HTTP request. For calculating the share of each part (network, execution of EdgeConnect and Flask application) thereto timestamps were logged during the code execution on the server at these events: request received, start of EdgeConnect subprocess, end of EdgeConnect subprocess and request fulfilled (see section \ref{cha:back_end} for reference). The measurements show that the execution of the Flask Server code only produces a small impact (3\% of the whole request duration). In addition, the share of the data transfer (network) is also small (5\%), as the size of the request and response payload is quite low in comparison to other modern-day web applications. The vast majority of the time needed to fulfill the HTTP request is caused by executing EdgeConnect (92\%). Overall, processing an inpainting takes overall around four seconds.

\section{Conclusion and Future Outlook}
As a result of this work, the Medical 3D Viewer on the website StudierFenster (\textit{http://studierfenster.tugraz.at/}) was extended by a tool for inpainting. All requirements to accomplish this task, like the integration of an inpainting module to the server-side of the website and creating a graphical user interface on the client-side, have been satisfied successfully. By using this tool, it is possible to use inpainting to remove the dissection from a dissected aorta in a CTA scan, directly in a web browser without installing any software. This functionality allows the creation of image pairs before and with an aortic dissection, which are in general unavailable from medical studies, by taking advantage of an intuitive graphical user interface. No knowledge about executing command-line scripts is needed to use the inpainting tool. In contrast to executing EdgeConnect directly via the command line, the mask creation is also simplified.

Looking at the results presented in Figure \ref{fig:result_}, one can see that the depicted dissected aorta is successfully changed to a healthy looking aorta. But it is also noticeable, that the region reconstructed by the inpainting is slightly blurred, which could be addressed by refinement of the EdgeConnect model. The timing analysis in section \ref{cha:results} shows that a call of the inpainting tool is completing in around four seconds, wherein the majority is caused by executing the neural network. This indicates the code of the inpainting tool itself offers little room for speeding up the application.

Concluding the current work on the tool, there are still several areas of improvement. Currently the tool only supports 2D drawings because only one masked slice is transferred to the server. On the client side the code basis for sending multiple slices to the server already exists, but it is deactivated at the moment. For 3D support this code section must be reviewed and in the back-end the inpainting route of the Flask application, and the model used for EdgeConnect has to be updated.

In future, user feedback could be utilized for the refinement of the model used for creating the edge map. Therefore, the user should be given the opportunity to revise the edge map created by the first stage of EdgeConnect. Subsequently, the corrections made by the user can be used for additional training of the model. Regarding the current functionality the adaption of the region of interest when rescaling the browser window is in an experimental stage at the moment. Hence, in future work this feature should be updated.

\section{Acknowledgments}
This work received funding from the TU Graz Lead Project “Mechanics, Modeling and Simulation of Aortic Dissection” and the Austrian Marshall Plan Foundation Scholarship 942121022222019. In addition, the Austrian Science Fund (FWF) KLI 678-B31. Further, this work was supported by CAMed (COMET K-Project 871132), which is funded by the Austrian Federal Ministry of Transport, Innovation and Technology (BMVIT), and the Austrian Federal Ministry for Digital and Economic Affairs (BMDW), and the Styrian Business Promotion Agency (SFG). Finally, we want to acknowledge the Overseas Visiting Scholars Program from the Shanghai Jiao Tong University (SJTU) in China. StudierFenster tutorial videos, also about the aortic dissection inpainting, can be found under our \textcolor{blue}{\href{https://www.youtube.com/channel/UCSe-q1nicDVwC550dngQT2w}{StudierFenster YouTube channel}}.

\bibliographystyle{plain}
\bibliography{smplbib}

\clearpage

\end{document}